\documentclass[letterpaper]{article} 
\usepackage{aaai25}  
\usepackage{times}  
\usepackage{helvet}  
\usepackage{courier}  
\usepackage[hyphens]{url}  
\usepackage{graphicx} 
\usepackage{natbib}  
\usepackage{caption} 
\usepackage{algorithm}
\usepackage{algorithmic}

\usepackage{newfloat}
\usepackage{listings}
\DeclareCaptionStyle{ruled}{labelfont=normalfont,labelsep=colon,strut=off} 
\floatstyle{ruled}
\newfloat{listing}{tb}{lst}{}
\floatname{listing}{Listing}
\newcommand{\bestoverall}[1]{\textbf{#1}}
\newcommand{\bestcolumn}[1]{\underline{#1}}

\title{Multiple Distribution Shift - Aerial (MDS-A): A Dataset for Test-Time Error Detection and Model Adaptation}

\author {
    Noel Ngu\textsuperscript{\rm 1},
    Aditya Taparia\textsuperscript{\rm 1},
    Gerardo I. Simari\textsuperscript{\rm 2},
    Mario Leiva\textsuperscript{\rm 2},
    Jack Corcoran\textsuperscript{\rm 3},\\
    Ransalu Senanayake\textsuperscript{\rm 1},
    Paulo Shakarian\textsuperscript{\rm 1},
    Nathaniel D. Bastian\textsuperscript{\rm 4}
}
\affiliations {
    \textsuperscript{\rm 1}Arizona State University, Tempe, AZ USA\\
   \textsuperscript{\rm 2}Department of Computer Science and Engineering, Universidad Nacional del Sur and Institute for Computer Science and Engineering, 
Bahía Blanca, Argentina\\
    \textsuperscript{\rm 3}U.S. Department of Defense, Arlington, VA USA\\
    \textsuperscript{\rm 4}United States Military Academy, West Point, NY USA\\
    nngu2@asu.edu, ataparia@asu.edu, gis@cs.uns.edu.ar, mario.leiva@cs.uns.edu.ar, jack.fd.corcoran@gmail.com, ransalu@asu.edu, pshak02@asu.edu,  nathaniel.bastian@westpoint.edu
}

\def\mdsawebsite{https://lab-v2.github.io/mdsa-dataset-website}
\begin{document}

\maketitle

\begin{abstract}
Machine learning models assume that training and test samples are drawn from the same distribution. As such, significant differences between training and test distributions often lead to degradations in performance. We introduce Multiple Distribution Shift - Aerial (MDS-A) - a collection of inter-related datasets of the same aerial domain that are perturbed in different ways to better characterize the effects of out-of-distribution performance.  Specifically, MDS-A is a set of simulated aerial datasets collected under different weather conditions. We include six datasets under different simulated weather conditions along with six baseline object-detection models, as well as several test datasets that are a mix of weather conditions that we show have significant differences from the training data. In this paper, we present characterizations of MDS-A, provide performance results for the baseline machine learning models (on both their specific training datasets and the test data), as well as results of the baselines after employing recent knowledge-engineering error-detection techniques (EDR) thought to improve out-of-distribution performance.  The dataset is available at \mdsawebsite.
\end{abstract}

%

\section{Introduction}
The robustness of models for object-detection remain a critical challenge when dealing with distributional shifts in real-world data. Distributional shifts in weather are especially important in aerial imagery since visibility and object-recognition can be heavily influenced by the weather.  Prior work on establishing benchmarks for out-of-distribution (OOD) object detection has largely focused on evaluating existing model performance during such a shift~\cite{Mao_Chen_Zhu_Chen_Su_Zhang_Xue_2023,Gardner_Popovic_Schmidt_2024}.  In this work, we present the Multiple Distribution Shift - Aerial (MDS-A) dataset - a collection of generated and labeled datasets with varying distribution differences and an associated set of baseline models.  To control experiments, we keep the baseline domain (aerial imagery) constant and perturb it in different ways to better characterize the effects of out-of-distribution performance. Specifically, MDS-A is a set of simulated aerial datasets taken under different weather conditions. We include six datasets under different simulated weather conditions along with six baseline object detection models as well as several test datasets that are a mix of weather conditions that we show have significant differences from the training data. In this paper, we present characterizations of MDS-A, provide performance results for the baseline models (on both in in-distribution and out-of-distribution test sets), as well as results of the baselines after employing recent knowledge-engineering error-detection techniques (error detection rules, or EDR~\cite{Kricheli_Vo_Datta_Ozgur_Shakarian_2024,Xi_Scaria_Bavikadi_Shakarian_2024,Lee_Ngu_Sahdev_Motaganahall_Chowdhury_Xi_Shakarian_2024,Shakarian_Simari_Bastian_2025}) thought to improve out-of-distribution performance. The rest of the paper is organized as follows.  First, we introduce the dataset, describing how it was constructed, and reporting on key statistics, importantly measures of distributional differences between the various training and testing sets.  Then, we describe how we trained a series of baseline models, and report on their performance both with and without error detection rules.  Finally, we discuss future research directions in the conclusion.



\section{Dataset}
In this section, we describe how we created the MDS-A dataset and report key statistics including measures of distributional differences.

\subsubsection{AirSim simulator}
\begin{figure*}[t]
\centering
\includegraphics[width=1\textwidth]{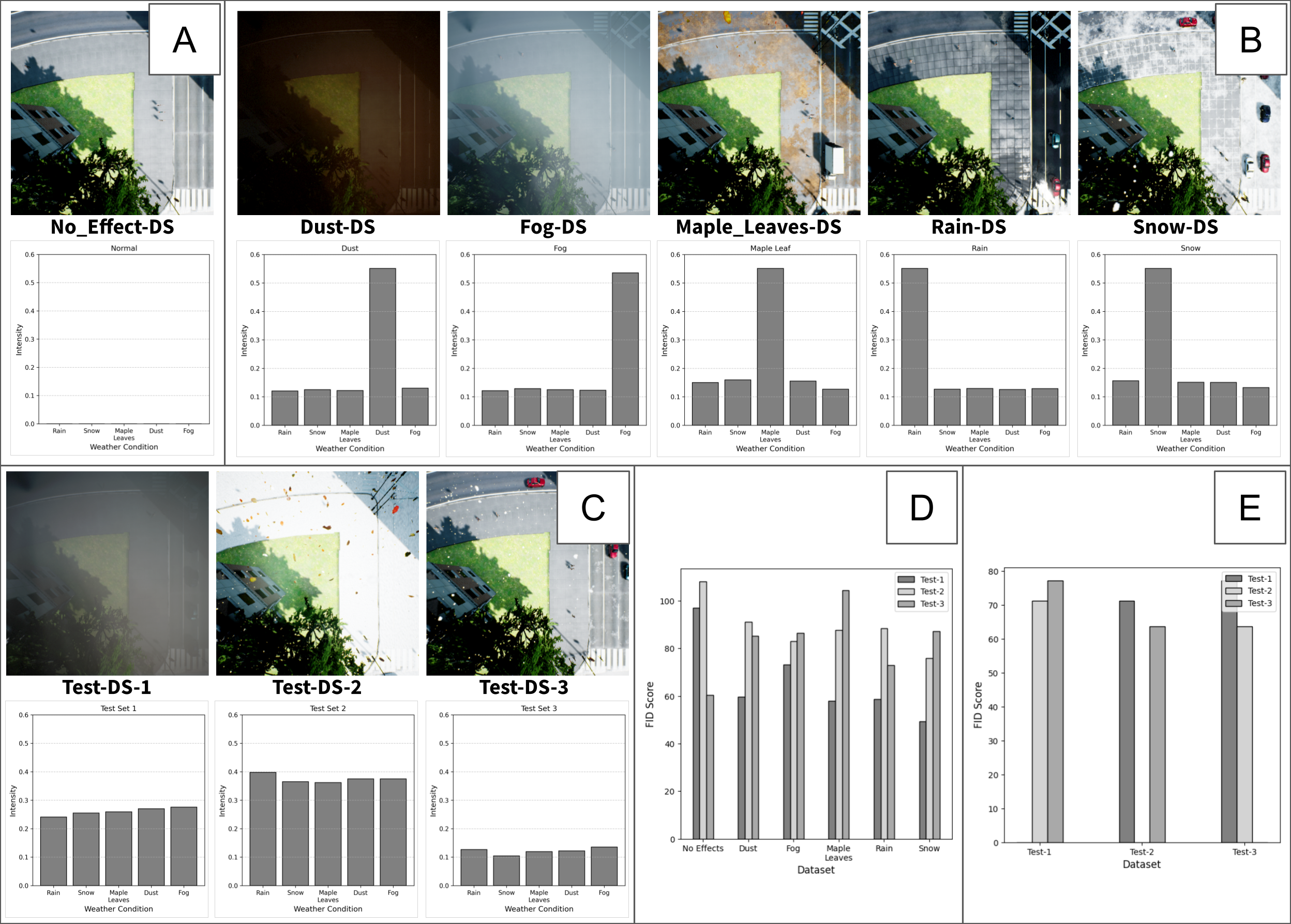} 
\caption{A) Image captured in AirSim with no weather effects applied along with a histogram showing the distribution of weather conditions of the dataset that it represents. B) Images captured in the same position in AirSim under different weather conditions: dust, fog, maple leaves, rain, snow-along with a histogram showing the distribution of weather conditions of the dataset that it represents. C) Images captured in the same position in AirSim with a mix of weather conditions applied along with a histogram showing the distribution of weather conditions of the dataset that it represents. D) A histogram showing the FID scores between the training sets and the 3 test sets. E) A histogram showing the FID score comparisions between the test sets.}
\label{fig:methodology_airsim}
\end{figure*}

To investigate the impact of distributional shift on aerial imagery in the context of weather conditions, we employed AirSim, an open-source simulator for drones, ground vehicles, cars, and other objects \cite{Shah_Dey_Lovett_Kapoor_2017}, to create a dataset of aerial imagery under various weather conditions. AirSim provides tools to capture images from different positions under different weather conditions by adjusting configurable parameters for effects such as dust, rain, fog, snow, and maple leaves. These parameters give us control over the intensity of various weather effects in the simulated scenes. Panel A and B in Figure \ref{fig:methodology_airsim} demonstrates how changing these parameters visually impact the images captured in AirSim.

\subsubsection{Data collection} 
For this study, the drone vehicle in AirSim was utilized to capture images from a top-down view at random positions within a simulated city environment. Given evidence that state-of-the-art object detection models are often susceptible to diverse weather conditions \cite{Pathiraja2024arxiv}, we configured AirSim with the following weather effects: rain, snow, fog, maple leaves, and dust. Using AirSim, images along with their bounding boxes were generated. Objects in the captured scenes were labeled by the research team to be classified into the following four categories: pedestrians, vehicles, nature, and construction. Each bounding box was assigned to exactly one of the categories. 

\begin{itemize}
    \item \textbf{Training sets} Multiple training sets were created, each focusing on a specific weather condition. For each dataset, the corresponding weather parameter for a weather condition (e.g.rain, snow, fog, maple leaves, or dust) are set randomly with a specific weather condition set to a particularly high value, while the other weather parameters were set to low values. As a result, we created distinct training datasets for the following conditions: Rain, Snow, Fog, Maple leaves, Dust. Panel B in Figure \ref{fig:methodology_airsim} shows the average intensities of each weather condition in each training set. In addition, a training set with no weather effects was created as well. The objective of these training sets are to enable the models to specialize in identifying objects under a single dominant weather condition.
    \item \textbf{Test sets} The test set was designed to evaluate the ability of models (trained on the training datasets) on a dataset that was created to simulate natural distributional shifts in weather. Unlike the training set, the test set consists of complex weather conditions where multiple weather conditions could be set to high values simultaneously, creating more challenging object-detection samples for the models. Panel C in Figure \ref{fig:methodology_airsim} shows the average intensities of each weather condition in the test set.
\end{itemize}

\subsubsection{Dataset Statistics} 
MDS-A consists of training sets that focus on a single weather condition, with each set containing 1000 images. The following training sets were generated: No-Effect Train Set, Dust Train Set, Fog Train Set, Maple-Leaves Train Set, Rain Train Set, Snow Train Set. The test sets, in contrast, feature a complex combination of weather conditions, also comprising 1000 images. 
Table \ref{table:dataset_statistics} shows some statistics regarding the number of images and the number of bounding boxes in each training set and test set.  
We note that with each of the six training sets, there is also a corresponding in-distribution hold-out set containing 100 images- this allows us to compare model in-distribution performance with out-of-distribution performance easily, all the while controlling for other factors.

\begin{table}
\centering
\begin{tabular}{|c|c|c|}
\hline
Name & Images & Bounding boxes \\
\hline
No-Effect Train Set & 1000 & 13320 \\
Dust Train Set & 1000 & 11257 \\
Fog Train Set & 1000 & 11099 \\
Maple-Leaves Train Set & 1000 & 12295 \\
Rain Train Set & 1000 & 11528 \\
Snow Train Set & 1000 & 11462 \\
\hline
No-Effect Test Set & 100 & 1267 \\
Dust Test Set & 100 & 1255 \\
Fog Test Set & 100 & 1406 \\
Maple-Leaves Test Set & 100 & 1077 \\
Rain Test Set & 100 & 1299 \\
Snow Test Set & 100 & 1368 \\
\hline
Test Set 1 & 1000 & 11117 \\
Test Set 2 & 1000 & 12466 \\
Test Set 3 & 1000 & 12558 \\
\hline
\end{tabular}
\caption{Statistics regarding the number of images and the number of bounding boxes in each training set and test set.}
\label{table:dataset_statistics}
\end{table}

Additionally, the Fréchet Inception Distance (FID) \cite{Heusel_Ramsauer_Unterthiner_Nessler_Hochreiter_2018} scores between the training sets and the test set are presented in Panel D in Figure \ref{fig:methodology_airsim}. These scores reflect the visual similarity between the training sets and the test set, providing a way to approximate the amount of distributional-shift between the training set and the test set. Higher FID scores, especially for conditions like Fog (73.3), suggests a larger distributional shift between the training set and the test set.

\subsubsection{Metadata Conditions} In addition to the datasets, we also provide additional meta conditions for each sample.  This information can be used to learn metacognitive models to identify potential errors. We use these in our baselines for error detection later in the paper. Examples of such conditions can be seen in Table \ref{tab:example_rls_Rules}

\begin{table}
\scriptsize
\centering
\begin{tabular}{|c|c|}
\hline
Rule & Meaning of Rule\\
\hline
$cond_{green}(w)$ &
Colors inside the bounding box has to be green \\

\hline 

$cond_{overlap}(w)$ & Pedestrians and vehicles should not overlap.
\\

\hline
\end{tabular}

\caption{Example EDCR Rule Learned for the MPSC Problem}
\label{tab:example_rls_Rules}
\end{table}

\section{Baseline Models and Associated Performance}
In addition to providing a dataset, we provide a series of baseline models, in addition to employing error detection rules \cite{Kricheli_Vo_Datta_Ozgur_Shakarian_2024,Xi_Scaria_Bavikadi_Shakarian_2024,Lee_Ngu_Sahdev_Motaganahall_Chowdhury_Xi_Shakarian_2024,Shakarian_Simari_Bastian_2025}.

\subsubsection{Model Training}
In order to establish a baseline for model performance under distributional shifts in the context of weather conditions, object-detection models were trained on each training set. The baseline object detection model that was used was DeTR \cite{Carion_Massa_Synnaeve_Usunier_Kirillov_Zagoruyko_2020} with a ResNet-50 \cite{He_Zhang_Ren_Sun_2016} backbone. 



The models were intentionally trained on a single training set without any mixes between training sets in order to emphasize different weather effects. These models were then evaluated on a more complex dataset aimed to emulate natural distributional shifts in weather conditions.

\subsubsection{In-Distribution Model Performance}
Table \ref{tab:training_distribution} provides results of the baseline models on their corresponding in-distribution dataset, specifically the \textit{No Effect Test Set}, \textit{Dust Test Set}, \textit{Fog Test Set}, \textit{Maple-Leaves Test Set}, \textit{Rain Test Set}, and \textit{Snow Test Set} (see statistics in Table~\ref{table:dataset_statistics} for details).  Here we report precision, recall, and F1 (harmonic mean of precision and recall). We note that model performance is generally consistent across the various models.

\subsubsection{Performance of Baseline Models on Test Sets}
The baseline models were evaluated on out-of-distribution test sets to assess their robustness under complex weather conditions that differ from the distribution in which they were trained - this is to establish a baseline for out-of-distribution performance on the three test sets in MDS-A. Table \ref{tab:model_performance_on_test_set_1} shows an expected decline in precision, recall, and F1 compared to in-distribution results. 

\begin{table*}
\begin{center}
\begin{tabular}{|c|c|c|c|c|c|c|}
\hline
& & & & Precision & Recall & F1 \\
Model & Precision & Recall & F1 & (EDR) & (EDR) & (EDR) \\
\hline
\multicolumn{7}{|c|}{Test Set 1} \\
\hline
No Effect Model & 0.35 & 0.27 & 0.31 & \bestoverall{\bestcolumn{0.62}} & 0.25 & 0.36 \\
Snow Model & 0.59 & \bestoverall{\bestcolumn{0.55}} & \bestoverall{\bestcolumn{0.57}} & 0.61 & 0.50 & 0.55 \\
    Dust Model & 0.59 & 0.54 & \bestoverall{\bestcolumn{0.57}} & 0.61 & 0.49 & 0.54 \\
Maple Leaf Model & \bestcolumn{0.60} & \bestoverall{\bestcolumn{0.55}} & \bestoverall{\bestcolumn{0.57}} & 0.60 & \bestoverall{\bestcolumn{0.55}} & \bestoverall{\bestcolumn{0.57}} \\
Rain Model & \bestcolumn{0.60} & 0.54 & \bestoverall{\bestcolumn{0.57}} & 0.60 & 0.54 & \bestoverall{\bestcolumn{0.57}} \\
Fog Model & 0.56 & 0.53 & 0.55 & 0.56 & 0.53 & 0.55 \\
\hline
\multicolumn{7}{|c|}{Test Set 2} \\
\hline
No Effect Model & 0.16 & 0.14 & 0.15 & \bestoverall{\bestcolumn{0.54}} & 0.13 & 0.21 \\
Snow Model & 0.44 & \bestoverall{\bestcolumn{0.26}} & \bestoverall{\bestcolumn{0.32}} & 0.47 & \bestcolumn{0.25} & \bestoverall{\bestcolumn{0.32}} \\
Dust Model & 0.43 & 0.25 & \bestoverall{\bestcolumn{0.32}} & 0.46 & 0.24 & \bestoverall{\bestcolumn{0.32}} \\
Maple Leaf Model & 0.45 & 0.25 & \bestoverall{\bestcolumn{0.32}} & 0.45 & \bestcolumn{0.25} & \bestoverall{\bestcolumn{0.32}} \\
Rain & \bestcolumn{0.46} & 0.25 & \bestoverall{\bestcolumn{0.32}} & 0.46 & \bestcolumn{0.25} & \bestoverall{\bestcolumn{0.32}}\\
fog & 0.40 & 0.25 & 0.31 & 0.40 & \bestcolumn{0.25} & 0.31 \\
\hline
\multicolumn{7}{|c|}{Test Set 3} \\
\hline
No Effect Model & 0.50 & 0.35 & 0.41 & \bestoverall{\bestcolumn{0.65}} & 0.30 & 0.41 \\
Snow Model & \bestcolumn{0.63} & 0.52 & \bestoverall{\bestcolumn{0.57}} & \bestoverall{\bestcolumn{0.65}} & 0.49 & 0.56 \\
Dust Model & 0.58 & 0.47 & 0.52 & 0.60 & 0.43 & 0.50 \\
Maple Leaf Model & 0.61 & \bestoverall{\bestcolumn{0.53}} & \bestoverall{\bestcolumn{0.57}} & 0.61 & \bestoverall{\bestcolumn{0.53}} & \bestoverall{\bestcolumn{0.57}} \\
Rain Model & 0.57 & 0.47 & 0.52 & 0.57 & 0.47 & 0.52 \\
Fog Model & 0.55 & 0.42 & 0.48 & 0.55 & 0.42 & 0.48 \\
\hline
\end{tabular}
\end{center}
\caption{Table showing the before and after results of applying EDR. Underlined numbers indicates the best model. Bold numbers indicates the best performing model across both baseline and EDR.}
\label{tab:model_performance_on_test_set_1}
\end{table*}

\subsubsection{Models with Error Detection Rules}
To enhance the robustness of the baseline models, error detection rule learning (EDR) was applied using the DetRuleLearn algorithm \cite{Xi_Scaria_Bavikadi_Shakarian_2024} with the hyperparameter of $\epsilon$ set to 0.5. Note that the rules were trained on the same data as the models.v  The application of EDR showed improvements in Precision while mostly maintaining F1 across all test sets as shown in Table 3.  This is due to the fact that EDR rules produce detections that are essentially recognizing that the model will most likely produce an error - and hence the results are discarded - resulting in a reduction of recall but an increase in precision.  We note that the results of \cite{Xi_Scaria_Bavikadi_Shakarian_2024} associate recall reduction with the $\epsilon$ hyperparameter (which would be up to an $0.5$ reduction, see Theorem 2 in \cite{Xi_Scaria_Bavikadi_Shakarian_2024} ) - however it is noteworthy that the reduction in recall is much less than predicted by the theoretical guarantee.

\begin{table}
\begin{center}
\begin{tabular}{|c|c|c|c|}
\hline
Model & Precision & Recall & F1 \\
\hline
No Effect & 0.75 & 0.62 & 0.68  \\
Snow &  0.75 & 0.69 & 0.72  \\
Dust & 0.75 & 0.65 & 0.70  \\
Maple Leaves & 0.76 & 0.70 & 0.73  \\
Rain & 0.75 & 0.65 & 0.70 \\
Fog & 0.73 & 0.62 & 0.67 \\
\hline
\end{tabular}
\end{center}
\caption{Table showing the performance of the baseline models trained on different training sets on an in-distribution dataset that is distinct from the training set.}
\label{tab:training_distribution}
\end{table}

\section{Conclusion and Future Work}
In this paper, we introduced the Multiple Distribution Shift - Aerial (MDS-A) dataset, a collection of simulated aerial datasets made to investigate the impact of distributional shifts, in the context of weather conditions, on object-detection model performance. Using the AirSim simulator, we created training datasets under six distinct weather conditions—rain, snow, fog, maple leaves, dust, and no effects—and evaluated the performance of baseline object-detection models trained on each condition using a complex test set that combines multiple weather effects. We also provide a suite of baseline models and in this paper we report on their performance for both in-distribution and out-of-distribution datasets.  Additionally, we also provide a baseline using error detection rules, which mitigates the degradation of precision. As we intend this to be a challenge dataset, we released MDS-A and the associated baseline models at \mdsawebsite.

Recent advances in topics such as test time training~\cite{Liang_He_Tan_2025}, domain generalization~\cite{Zhou_Liu_Qiao_Xiang_Loy_2023}, and meta learning~\cite{Vanschoren_2018} are all potential candidates for improving performance. Further, this dataset allows the exploration of novel ensemble methods based on models trained on different distributions.

\section{Acknowledgments}
This research was supported by the Defense Advanced Research Projects Agency (DARPA) under Cooperative Agreement No. HR00112420370, the U.S. Army Combat Capabilities Development Command (DEVCOM) Army Research Office under Grant No. W911NF-24-1-0007, and the U.S. Army DEVCOM Army Research Lab under Support Agreement No. USMA 21050. The views expressed in this paper are those of the authors and do not reflect the official policy or position of the U.S. Military Academy, the U.S. Army, the U.S. Department of Defense, or the U.S. Government.

\bigskip
\noindent Thank you for reading these instructions carefully. We look forward to receiving your electronic files!

\bibliography{aaai25}

\end{document}